# The Agent-based Modelling for Human Behaviour Special Issue

Soo Ling Lim [1], Peter Bentley [1,2]

**Corresponding:** Soo Ling Lim (s.lim@cs.ucl.ac.uk)

1. Department of Computer Science, University College London, London, UK
2. Autodesk Research, London, UK

"We inhabit an obscure planet, in an obscure galaxy, around an obscure sun, but on the other hand, modern human society represents one of the most complex things we know." – David Christian (Christian, 2014).

If human societies are so complex, then how can we hope to understand them? Artificial Life gives us one answer. The field of Artificial Life comprises a diverse set of introspective studies that largely ask the same questions, albeit from many different perspectives: Why are we here? Who are we? Why do we behave as we do? Starting with the origins of life provides us with fascinating answers to some of these questions. However, some researchers choose to bring their studies closer to the present day. We are after all, human. It has been a few billion years since our ancestors were self-replicating molecules. Thus, more direct studies of ourselves and our human societies can reveal truths that may lead to practical knowledge.

The papers in this special issue bring together scientists who choose to perform this kind of research. Expanded from submissions to our annual Agent-Based Modelling of Human Behaviour Workshop, the studies share similar methods, all using variations of agent-based-modelling (ABM) to ask their own "what if" questions. As guest editors, we believe such collections help bring together and enhance such research by sharing ideas. While ABM research – out of necessity – is often highly specialised towards the hypotheses and phenomena under study, the research methodology is shared by all (Macal, 2016). We formulate our hypothesis, develop our agent-based model of the relevant aspects of reality, and run experiments to gather evidence which may support or refute the hypothesis. An experimental model that supports the hypothesis may not prove that reality follows this approach or agrees with this result; but it indicates that there exist a specific set of conditions which if found to be true elsewhere may produce the same result. Modelling tells us about trends, about possible likelihoods. Our ABMs show us what will result if our assumptions are valid and why, whether we are examining civil violence (Quek et al., 2009), app stores (Lim et al., 2016), the economy (Farmer and Foley, 2009), fish markets (Kirman & Vriend, 2001), language evolution (Griffiths & Kalish, 2007), or energy consumption (Bentley et al., 2021). When we study human societies, ABMs are the tools of choice for obvious reasons: it is not ethical or safe to play "what if" experiments with ourselves. The researchers in this special issue demonstrate the exciting potential in ABM. We can create our own safe virtual worlds and make discoveries that enlighten us about ourselves.

In this special issue we have a fascinating collection of studies that relate to us. For example, from our first paper, "The effects of information on the formation of migration routes and the dynamics of migration" by Hinsch and Bijak we may discover that as we spread across the planet both physically and culturally, the nature of our interactions affects ourselves in ways we may not expect. The authors developed a spatially explicit agent-based model of human migration and showed that information exchange plays a crucial role in affecting all aspects of migration: too little information results in suboptimal migration routes, increased stochasticity and migrants frequently not arriving at their preferred destinations, and too much information under certain conditions can lead to less predictable migration routes.

Likewise, from our second paper "Adapting the exploration-exploitation balance in heterogeneous swarms: tracking evasive targets" by Kwa et al. we might recognise that like the robotic agents, we are diverse in every sense of the word, and that diversity is not only to be celebrated, it is a necessity for the optimal functioning of our societies. In their work, agents with different properties and behaviours are part of the same collective, forming a heterogeneous swarm. The authors used a decentralized search and tracking strategy with adjustable levels of exploration and exploitation and found that small heterogenous swarms can match and outperform homogeneous swarms. They also explored using differentiated strategies to take advantage of heterogenous swarms.

Our third contribution, "Expertise, social influence and knowledge aggregation in distributed information processing" by Mertzani et al. helps us understand that fields such as psychology provide insights into human behaviour and how we communicate – a topic close to our hearts as guest editors (Lim & Bentley, 2018, Lim & Bentley, 2019a, Lim & Bentley, 2019b, Schoots et al., 2019, Guo et al., 2020, Lim et al., 2023). Mertzani et al. developed an algorithm for knowledge aggregation based on Nowak's Regulatory Theory of Social Influence. This theory posits that social influence consists not only of sources trying to influence targets, but also targets seeking sources by whom to be influenced, and learning what processing rules those sources are using.

Our societies face constant challenges that threaten us. Sadly, humans evolved to be violent, and to work together for our violent aims, as ongoing geopolitical instability shows all too well. "Explaining neuro-evolution of fighting creatures through virtual fMRI" by Godin-Dubois et al. investigates where these behaviours come from and what causes different strategies and morphologies. They developed artificial creatures that engage in individual and team competitions and investigated three types of reactions: pain, vision and audition each with varying response levels but consistent across all evolution types.

When we're not battling each other, human societies also must battle pathogens – most recently in the global COVID-19 pandemic. As the virus becomes endemic and we develop coping strategies, are we responding appropriately to the changing risk levels? "Self-isolation and testing behaviour during the COVID-19 pandemic: an agent-based model" by Gostoli and Silverman models this phenomenon to find out. In this paper the authors developed an agent-based model which includes a behavioural module determining the agents' testing and isolation propensity. According to them, most models focus on the replication of the interactions' processes through which the virus is passed on from infected agents to susceptible ones; in this model, agents modify their behaviour as they adapt to the risks posed by the pandemic.

Finally, there can be no more fundamental challenge to a society than its constant search for resources to sustain itself. Our final paper "Social search and resource clustering as emergent stable states" by Luthra and Todd looks at the interplay between resources and consumers of those resources in a 2D simulated world. Agents evolve strategies to search for food and the resources resemble plants which grow continuously across the two-dimensional simulated world. Resources and other consumers produce distinguishable "odours" that are normally distributed around their position in the grid and agents can perceive these two types of odours and make movement decisions that take them toward or away from each kind of source.

Through models such as these can learn why we might behave as we do as a species, and whether we can start to mitigate or redirect some of the less-than-optimal patterns of behaviours that our complex societies might be attracted towards.

We hope you enjoy reading these articles as much as we have, and we hope that you might be inspired to ask your own "what if" questions using ABM in your work.